\PassOptionsToPackage{pdftex,dvipsnames}{xcolor}
\PassOptionsToPackage{hidelinks}{hyperref}
\documentclass[sigconf]{acmart}
\usepackage{booktabs} 
\usepackage{tabularx}

\usepackage{amsmath}
\usepackage{amssymb}
\usepackage{bm}
\usepackage{wrapfig}
\usepackage{multirow}
\usepackage{multicol}
\usepackage{enumitem}
\usepackage{array}
\usepackage{color, soul}
\usepackage[dvipsnames]{xcolor}

\usepackage{setspace}

\usepackage{subcaption}

\usepackage{graphicx,array}
\graphicspath{ {fig/} }

\newcommand{\I}[1]{\textit{#1}}

\usepackage{xargs}
\usepackage[colorinlistoftodos,prependcaption,textsize=tiny]{todonotes}
\newcommandx{\unsure}[2][1=]{\todo[linecolor=red,backgroundcolor=red!25,bordercolor=red,#1]{#2}}
\newcommandx{\change}[2][1=]{\todo[linecolor=blue,backgroundcolor=blue!25,bordercolor=blue,#1]{#2}}
\newcommandx{\info}[2][1=]{\todo[linecolor=OliveGreen,backgroundcolor=OliveGreen!25,bordercolor=OliveGreen,#1]{#2}}
\newcommandx{\improvement}[2][1=]{\todo[linecolor=Plum,backgroundcolor=Plum!25,bordercolor=Plum,#1]{#2}}
\newcommandx{\thiswillnotshow}[2][1=]{\todo[disable,#1]{#2}}
\setcopyright{acmcopyright}

\copyrightyear{2017}
\acmYear{2017}
\setcopyright{acmcopyright}
\acmConference{ACM-BCB'17}{}{August 20-23, 2017, Boston, MA, USA.}
\acmPrice{15.00}
\acmDOI{http://dx.doi.org/10.1145/3107411.3107485}
\acmISBN{978-1-4503-4722-8/17/08}

\begin{document}

\title{Identifying Harm Events in Clinical Care \\ through Medical Narratives}

\author{Arman Cohan}
\affiliation{%
\institution{Information Retrieval Lab, Dept. of Computer Science\\
 Georgetown University}
}
\email{arman@ir.cs.georgetown.edu}

\author{Allan Fong}
\affiliation{%
  \institution{National Center for Human Factors in Healthcare\\
  MedStar Health}
}
\email{allan.fong@medstar.net}

\author{Raj Ratwani}
\affiliation{%
  \institution{National Center for Human Factors in Healthcare\\
  MedStar Health}
}
\email{raj.m.ratwani@medstar.net}

\author{Nazli Goharian}
\affiliation{%
  \institution{Information Retrieval Lab, Dept. of Computer Science\\
   Georgetown University}
}
\email{nazli@ir.cs.georgetown.edu}

\newcommand\timex{\textsc{Timex3}}
\newcommand\event{\textsc{Event}}
\newcommand\B[1]{\textbf{#1}}
\newcommand\SC[1]{\textsc{#1}}

\newcommand*{\matr}[1]{\mathbf{#1}}
\newcommand*{\vect}[1]{\bm{#1}}

\begin{abstract}
Preventable medical errors are estimated to be among the leading causes of injury and death in the United States. To prevent such errors, healthcare systems have implemented patient safety and incident reporting systems. These systems enable clinicians to report unsafe conditions and cases where patients have been harmed due to errors in medical care. These reports are narratives in natural language and while they provide detailed information about the situation, it is non-trivial to perform large scale analysis for identifying common causes of errors and harm to the patients. In this work, we present a method based on attentive convolutional and recurrent networks for identifying harm events in patient care and categorize the harm based on its severity level. We demonstrate that our methods can significantly improve the performance over existing methods in identifying harm in clinical care.
\end{abstract}

%
%


\maketitle
\pagenumbering{gobble}


\section{Introduction}
\label{sec:intro}

Preventable medical errors have been shown to be a major cause of injury and death in the United States \cite{donaldson2000err,Williams2016,makary2016medical}. Medical errors are estimated to be the 3rd leading causes of death in the U.S. \cite{starfield2000us,makary2016medical}, which translates to an estimated incidence of 210,000 to 400,000 deaths annually \cite{James2013,makary2016medical}.
To address these major concerns, healthcare systems have adopted reporting systems in clinical care to help track and trend hazards and errors in patient care \cite{Mitchell2016,Williams2016}. The data from these systems are later used to identify the causes of harm and actions that should be taken to prevent similar situations. These reporting systems allow frontline clinicians to report events that are relevant to patient care including both near misses and serious safety events. Near misses are events or situations where a hazard was identified before a patient could be harmed. For example, a wrong medication order that was never administered to a patient would be considered a near miss, hence reported as a no-harm event. Serious safety events on the other hand are situations where a patient was harmed. The above example would be considered a patient harm event if the nurse had actually administered the medication causing additional treatment, monitoring, or irreversible effects on the patient.

Although reporting systems have been implemented with the goal of improving patient safety and patient care, hospital staff are faced with many challenges in analyzing and understanding these reports \cite{Macrae2016,Mitchell2016}. These reports which are narratives in natural language are generated by frontline staff and vary widely in content, structure, language used, and style.
These reports include a textual field where the clinicians describe the safety event and its details in free-form text. While these texts provide valuable information about the safety event, it is challenging to perform large scale analysis of these narratives to identify important safety events. In this paper, we propose and evaluate Natural Language Processing (NLP) methods to identify cases that caused harm to the patient based on medical narratives.

One important aspect in patient care is to identify events that have contributed to or resulted in harm to the patient \cite{Williams2016}. There have been many efforts in characterizing harm to the patients based on their severity. The most common is a harm categorical system that indicates the severity of the harm to the patient \cite{harm2012categories}. These categories range from an unsafe condition (which describes an event where there was no error, but had capacity to cause harm) to death (which is an event where an error has caused or contributed to the death of a patient). These harm categories are described in Table \ref{tab:cat} in detail \cite{harm2012categories}.

\begin{table*}[tb]
\centering
\footnotesize
\renewcommand{\arraystretch}{1.1}
\caption{Categories of Errors in Patient Care as defined by Agency for Healthcare Research and Quality \cite{harm2012categories}. The severity of harm increases from top to bottom. Categories \{E,F,G,H,I\} are harm categories while \{A,B1,B2,C,D\} are no-harm events.}
\label{tab:cat}
\begin{tabular}{@{}lp{0.47\textwidth}p{0.45\textwidth}@{}}
\toprule
Cat. & Description                                                                                                & Example                                                                                      \\ \midrule
A        & No error, capacity to cause error                                                                          & Confusing equipment                                                                                           \\
B1        & Error that did not reach the patient (due to chance)                                                                      & Wrong medication label discovered                                                                                           \\
B2        & Error that did not reach the patient (because of active recovery efforts by caregivers)                                                                      & Mislabeled specimen in a laboratory discovered on a regular checking                                                                                           \\
C        & Error that reached patient but unlikely to cause harm             & Multivitamin was not ordered on admission                                                    \\
D        & Error that reached the patient and could have necessitated monitoring and/or intervention to preclude harm & Regular release metoprolol was ordered for patient instead of extended-release               \\
E        & Error that contributed to or resulted in temporary harm                                                                & Blood pressure medication was inadvertently omitted from the orders                          \\
F        & Error that could have caused temporary harm requiring initial or prolonged hospitalization                 & Anticoagulant, such as warfarin, was ordered daily when the patient takes it every other day \\
G        & Error that resulted in permanent harm                                                           & Immunosuppressant medication was unintentionally ordered at wrong dose              \\
H        & Error that necessitated intervention to sustain life                                            & Anticonvulsant therapy was inadvertently omitted                                             \\
I        & Error that contributed to or resulted in death                                                             & Beta-blocker was not reordered post-operatively                                              \\ \bottomrule
\end{tabular}
\end{table*}


The patient incident narratives can be complex and it is challenging to identify the cases of harm from these reports. These reports often consist of multiple events.
 For example, consider a case where a patient is found on the floor in the emergency department (ED) with no physical signs of injury. This is initially entered as a no-harm case. However, later when the patient is transferred to the radiology for an x-ray as a precaution, a small fracture is discovered from the x-ray. Therefore, while the ED staff originally entered the event as a no-harm event, the radiology department would revise this as a harm event.

For these reasons, reporting harm is often miscategorized. While most events are eventually recategorized by a department manager or patient safety officers who have a more global perspective of events, this recategorization incurs additional time, resources, and expenses leading to missed opportunities to address the actual event in a timely fashion. We present a method for identifying the severity of harm from narratives regarding incidents in patient care. While there is a growing number of work in categorizing patient safety reports, none has looked at the modeling of general harm across all event types \cite{ong2012automated,fong2015exploring}. Our method is based on a neural network model consisting of several layers including a convolutional layer, a recurrent layer, and an attention mechanism to improve the performance of the recurrent layer. Our method is designed to capture local significant features as well as the interactions and dependencies between the features in long sequences. Traditional methods in general and domain specific NLP rely heavily on engineering a set of representative features for the task and utilizing external knowledge and resources. While these models have been shown to work reasonably well for different tasks, their success relies on the type of features that they utilize. Apart from the feature engineering efforts, these approaches usually model the problem with respect to certain selected features and ignore other indicators and signals that might improve prediction.
In contrast, our approach only relies on the text in the patient incident narratives and it does not rely on any features or external resources, making it generalizable. Through extensive evaluation on two large datasets, we show that our proposed method is able to significantly outperform the existing approaches of identifying harm in clinical care. Effective identification of harm can help the hospital staff save time both during analysis and reporting. Furthermore, a more accurate and immediate classification of harm can also help to better prioritize resources to address safety incidents, which subsequently improves general patient care.

%

\section{Related Work}
\label{sec:related}
There has been a growing number of work in categorizing patient incident and safety narratives in clinical care. \citet{fong2015exploring} explored both the unstructured free-text and structured data elements in safety reports to identify and rank similar events. They evaluated different search methods utilizing bag of words features, structured elements, and topic modeling features to rank and identify similar events. In another work, \citet{ong2012automated} explored the similar problem of identifying extreme-risk events in patient safety and incident reports using Naive Bayes and SVM classifiers with bag of words features. In contrast to these works, we focus on the problem of identifying harm and categorizing the harm based on its severity in medical narratives. We present neural network methods that are able to capture information from the complex narratives regarding safety events without utilizing any external features. We compare our results with feature based methods and show that our proposed methods are significantly superior in identifying and classifying harm in patient incident reports.

The problem of identifying harm in patient safety reports is a type of text classification problem. Traditional approaches in text classification include methods to extract features from text and then use the feature vector as an input to a classifier such as SVM \cite{aggarwal2012survey}. More recently, neural networks have shown success in many NLP tasks including text classification. Two of the more widely used neural network architectures have been Convolutional Neural Networks (CNN) \cite{lecun1998gradient} and Recurrent Neural Networks (RNN) \cite{elman1990finding}. \citet{Collobert:2011} were one of the first to utilize CNNs in many NLP tasks including text classification. In particular, they proposed a CNN architecture which operated on one-hot encodings of words; their model was based on the original CNN architecture of \citet{lecun1998gradient} with adaptations to the NLP domain and showed improvements on several NLP tasks. Later CNNs were further explored for sentence modeling and classification tasks \cite{kalchbrenner:2014,kim:2014,dos2014deep}.

In the biomedical domain, there have been many efforts in classifying biomedical text and narratives based on different tasks. Many works have looked at the specific problem of indexing biomedical literature using MeSH\footnote{Medical Subject Heading} terms. These works have mostly used supervised learning frameworks with bag of words features, named-entities, and ontology specific features \cite{yepes2013comparison}. More recently, \citet{rios2015convolutional} utilized CNNs for this task; they showed that CNNs are more effective than feature based methods in biomedical indexing. The authors in \cite{Cohan2017} used a neural network architecture using CNN and RNNs to classify patient safety report. \citet{xu2016text} used a CNN architecture, with multiple sources of word embeddings and evaluated its effectiveness on the tasks of biomedical literature indexing and clinical note annotation. Our method, in contrast, is based on an extension of CNNs with recurrent layer as well as an attention model to improve performance on longer sequences. We compare our methods with a CNN baseline and show that our methods can significantly outperform the baselines. Our focus is on the challenging task of identifying harm in patient incident reports where the incident narratives are often complex, consisting of multiple chained events in a single narrative. Our proposed model, is designed to capture these complexities.

%


\section{Methods}
\label{sec:method}
We present a general neural network architecture for identifying harm in patient safety reports. As explained in \S \ref{sec:intro}, the narratives regarding patient safety can be complex and identifying harm to the patient is challenging in these reports. To be able to perform this task effectively, we will need a model that is able to capture both local features as well as the language usage in the entire report. To achieve this goal, we propose a neural network consisting of several layers where each layer is designed to address the aforementioned challenges. Our approach does not require feature engineering and it learns to identify significant features from the raw text automatically. We first describe the general outline of our model and then we describe each component in more detail.

\begin{figure}[t]
\begin{center}
\includegraphics[scale=0.71]{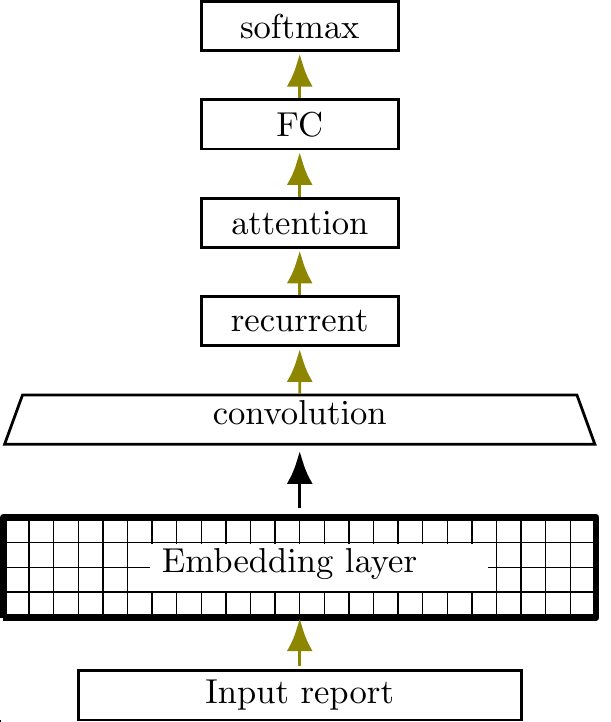}
\caption{The outline of the proposed model. Input report consists of the raw text of the report, embedding layer does the pre-processing and represents the input as a matrix. FC stands for Fully Connected layers.}
\label{fig:arch}
\end{center}
\end{figure}

\paragraph{The outline of the model} The proposed architecture is shown in the Figure \ref{fig:arch}. The input report is first pre-processed and represented as a matrix corresponding to word embeddings. Word emdeddings or distributed representations of words aim to embed (represent) words with dense vectors such that words with similar properties have similar vectors \cite{bengio2013representation}. These embeddings can be general and pre-trained or can be trained according to the task at hand. Then a convolutional layer extracts the significant local features that are helpful for identifying harm in the report. Next, a recurrent layer captures the interactions of the local features along the entire sequence of the words in the report. In the next layer, we propose an attention model which serves to overcome the problem of recurrent networks in compressing an entire sequence in a single vector by focusing the attention to the important timesteps (steps of the sequence) in the recurrent layer. Finally, the output of the attention model is a vector which is passed to a fully connected layer and a softmax classifier identifies the level of harm associated with the report. We now explain each of these layers in detail.

\subsection{Embedding layer}

This layer pre-processes the raw text corresponding to the medical report and represents it as a matrix of real valued numbers. This matrix consists of embeddings of the words in the report. We tokenize the text using a simple white space tokenizer and we lowercase all the words. We then transform the input sequence of tokens into a sequence of dense distributional vectors. Specifically, given a sequence of tokens $W$ where $W=\langle w_1, w_2, ..., w_n\rangle$ and $w_i$'s are the input sequence tokens, the embedding layer represents each token $w_i$ as a $d$ dimensional vector $x_i$, and the sequence $W$ will be represented as a matrix of real valued numbers $\matr{X}$ with dimensions of $\matr{X}\in\mathbb{R}^{(n_{\max}\times d)}$ where $n_{\max}$ is the maximum sequence length. Text inputs with length larger than $n_{\max}$ will be cropped and text inputs shorter than the $n_{\max}$ are padded with zeros. The value of $n_{\max}$ is determined empirically.

\begin{figure}[t]
\begin{center}
\includegraphics[scale=0.91]{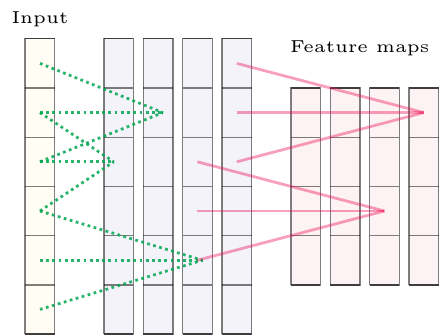}
\caption{A convolutional layer takes a series of tokens as input and applies $l$ filters of size $k$ (green dotted arrows in the figure) to derive the feature values over a local window of tokens; $l = 4$ and $k=3$ are shown here (all filters are the same size). To produce the component's output, a max pooling layer (red arrows in the figure) considers region sequences of length $n$ and keeps the highest feature value for the sequence $n = 3$ is shown here.}
\label{fig:conv}
\end{center}
\end{figure}

\subsection{Convolutional layer}

Convolutional layer is repsonsible for extracting local features from the input text. Convolutional Neural Networks (CNNs) \cite{lecun1998gradient} have been previously used in sentence modeling and classification tasks \cite{kalchbrenner:2014,kim:2014}. A CNN is a neural network that consists of two main operations: convolution and pooling. A convolution is an operation between two functions $f$ and $g$ where $f$ is the primary vector and $g$ is the filter. The convolution operation between $f$ and $g$, evaluated at entry $n$ is represented as:
$  (f*g)[n]=\sum_{i=-K}^{K} f[n-i]\times g[i]$

\noindent Where $*$ denotes the convolution operation and $L=2K-1$ is the length of the filter. Here, $f$ is the input to the convolution (word vectors obtained from the embedding layer).

Features are extracted by convolution of the input text with a number of linear filters, adding a bias term and applying a non-linearity. The result is called a feature map. The trained weights in these filters correspond to a linguistic feature detector that learns to recognize a specific class of n-grams where $L\le n$. A max-pooling operation is used after the convolution to extract the significant features.

\paragraph{Multiple feature maps}
Similar to convolutional networks for object recognition \cite{lecun1998gradient}, we use multiple feature maps with different filters to capture various aspects of the input sequence. Figure \ref{fig:conv} illustrates how the feature maps are constructed from the input. First the convolution and non-linearity are applied to the input and then the max pooling derives the resulting feature maps. The final output of this layer at each time-step is the concatenation of the feature maps at that time-step.

\subsection{Recurrent layer}

The result of the convolution layer is a sequence of vectors each of which is the concatenation of the feature maps at corresponding time-step. Convolutional layer is able to extract significant local features that are important for our task. However, the interactions between the words are not captured specially if the words are distant from each others. Recurrent Neural Networks (RNNs) are a family of neural networks that are designed to process a sequence of values. We use an RNN on top of the result of the convolution layer to capture interactions along the entire sequence of words.
RNNs are an extension of multilayer perceptrons in which the output of each step is used as an additional input to the next step. Specifically, the activations arrive at the hidden layer of the network from both the current external input and the hidden layer activations one step back in time. The general formulation of an RNN is as follows:
\begin{subequations}
\setstretch{0.7}
\begin{align}
  h(t) = g(W^{(h)} h(t-1) + W^{(x)} x(t)) \label{eq:rnn}\\
  \hat{y}(t) = \mathrm{softmax}(W^{(s)} h(t)) \
\end{align}
\end{subequations}

Where $h(t)$ shows the hidden state of the RNN in time step $t$, $x(t)$ is the input sequence at time step $t$, $W^{(h)}$, $W^{(x)}$, and $W^{(s)}$ are the weights associated with the hidden state, input, and softmax, respectively, and $g$ is an activation function such as RELU \cite{dahl2013improving}.

\begin{figure}[t]
\begin{center}
\includegraphics[scale=1]{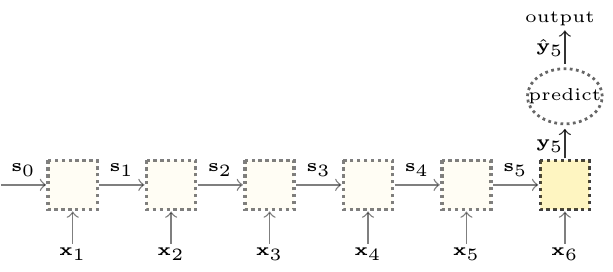}
\caption{Representing a sequence with an RNN and using it's last hidden state for making predictions.}
\label{fig:rnn}
\end{center}
\end{figure}

In sequence modeling tasks, the final hidden state of the network can represent the whole sequence and can be used for making predictions \cite{goldberg2016primer}. This final hidden state, in theory, can capture all the information in the entire sequence. This is because the output of each timestep is used as an input to the subsequent timestep in the network. Figure \ref{fig:rnn} illustrates the prediction made at the last hidden state of the network.

\subsubsection{RNN variants}
Training the general formulation of RNNs in practice is difficult due to the exploding and vanishing gradient problems (gradients becoming exceedingly high or become exceedingly close to 0 after only a few timesteps) \cite{pascanu2013difficulty}. For the exploding gradient problem, a common solution is to cap the gradient value at a specific maximum threshold. There has been some variants of RNNs that assist the gradient flow and mitigate the vanishing gradient problem. Most notable are the Long Short Term Memory (LSTM) \cite{hochreiter1997long} and Gated Recurrent Unit (GRU) \cite{cho2014properties}.

LSTM adds additional gates to the regular hidden layer of a recurrent network to assist the gradient flow and allow the network to be effectively trained. These gates control the amount of information to be forgotten or preserved throughout the sequence. Concretely, we are referring to the formulation of \citet{graves2013generating} for LSTM.

%

GRU proposed by \citet{cho2014properties} makes each recurrent unit to adaptively capture dependencies of different time scales and similar to LSTM, GRU also has gating units that control the flow of information through the computational graph. The difference with LSTM is that the GRU does not have a separate memory cell. We use the exact formulation of \citet{cho2014properties} for GRU.

%

\paragraph{Bidirectional RNNs}

In order to also capture the backward dependencies and interactions between different parts of a sequence, a backward RNN is also trained which can encode the information from the future time steps \cite{schuster1997bidirectional,graves2008}. The hidden states of the backward RNN are then considered along with the corresponding hidden states of the forward RNN (e.g. by concatenation) at each time step and used in the subsequent layers.

Let $\vect{x}=\langle x_1,...,x_n \rangle$ be the input to the Recurrent layer. Then the bidirectional RNN over the time steps $t=1,...,n$ is as follows:

\begin{equation}
h_t = \langle \overrightarrow{h_t};\overleftarrow{h_t} \rangle
\end{equation}

where ``$\langle \cdot;\cdot \rangle$'' shows the concatenation operation and $\overrightarrow{h_t}$ ($\overleftarrow{h_t}$) is the forward (backward) RNN defined as follows:

\begin{equation}
\overrightarrow{h_t}=\overrightarrow{RNN}(x_t); \;\;\; \overleftarrow{h_t}=\overleftarrow{RNN}(x_t)
\end{equation}

Where $\mathrm{RNN}(\cdot)$ is the feed forward RNN cell in the general form, LSTM, or GRU.

\subsection{Attention model}

We use an attention model on top of our recurrent layer to be able to capture the local features that are more important in the task at hand. The limitation of using the regular recurrent network for the classification task is that the last time step of recurrent network loses some information about the sequence, specially when the sequence length becomes large \cite{cho2014properties}. This will not be a significant problem in short sentence classification tasks, but in our problem, the reports can have several sentences and the sequence length can be long. While in theory, the last step of the RNN is able to encode all the important information in the entire sequence, in practice it tends to focus more on the more recent time steps \cite{sutskever2014sequence} and therefore loses some information specially about the earlier time steps. Using a bidirectional RNN can partially mitigate this problem where the last state of the backward RNN along with the last state of the forward RNN are able to capture the information in beginning and the end of the sequence. However, bidirectional RNNs are still suffering from the same information loss problem.

\begin{figure}[t]
\begin{center}
\includegraphics[scale=0.6]{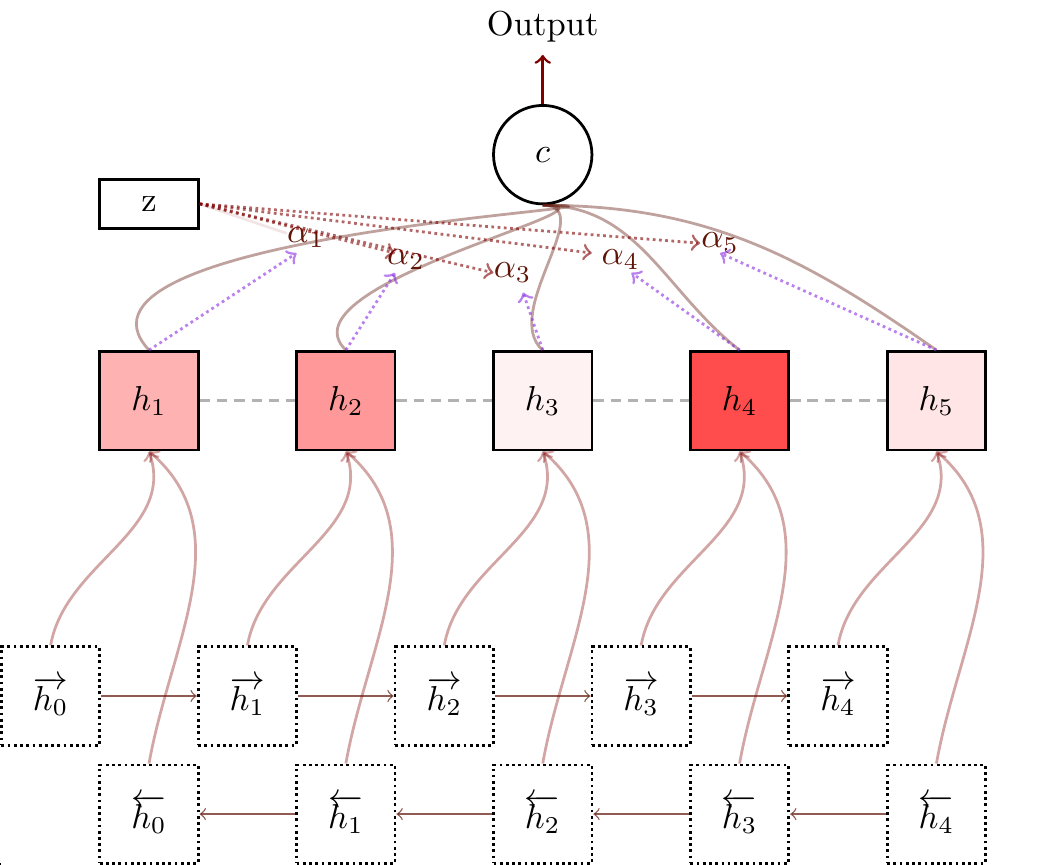}
\caption{The attention mechanism over the recurrent layer. The states $\protect\overrightarrow{h_i}$ and $\protect\overleftarrow{h_i}$ show forward and backward hidden state of the RNN respectively. $h_i$'s are the concatenation of the forward and backward RNN states. $z$ is a context vector that attends to important time steps and $\alpha_i$ are the weights associated with each hidden state $h_i$. The figure shows an example where darker colors for the $h_i$ show more importance in constructing $c$ which is used as input to the next layer.\label{fig:att}}
\end{center}
\end{figure}

Inspired by recent work in machine translation \cite{bahdanau2014neural} and document modeling \cite{yang-EtAl:2016:N16-13}, we propose to address this problem using a soft attention mechanism. Attention mechanism helps in constructing a context vector over the input that automatically incorporates the important parts of the input. The attention mechanism is shown in Figure \ref{fig:att}. Particularly, instead of only considering the last hidden state of the RNN ($h_n$), the attention model attends to the important timesteps by introducing additional weights ($\alpha$'s):
\begin{equation}
c=\sum\limits_{t=1}^N \alpha_t h_t
\end{equation}

\noindent where $t$ are the time steps in the input sequence and the weights $\alpha$ are learned according to the following softmax function:
\begin{equation}
  \alpha_t = \mathrm{softmax}(u_t^\intercal z) \label{eq:att}
\end{equation}

\noindent where $z$ is a context vector that helps in finding the weight importance of the local states $h_t$. This context vector can be seen as an input memory representation in memory networks and is jointly trained with the network.
$u_t$ is a feed forward function which we will define later and for a set of scores $s_i$, the softmax function returns a probability distribution over the scores:
\begin{equation}
  \mathrm{softmax}(s_i)=\frac{\exp(s_i/\beta)}{\sum_j\exp(s_j/\beta)};
\end{equation}

\noindent with $\beta$ being a parameter controlling the smoothness of the resulting distribution.

$u_t$ in equation \ref{eq:att} is the result of applying a regular feed forward network over the hidden state $h_t$ with weights $U$ and biases $b$:

\begin{equation}
  u_t = F(h_t) = \tanh(Uh_t + b)
\end{equation}

Finally, our model at the top layer has a fully connected layer followed by a softmax non-linear layer that predicts the probability distribution of harm severity given an input report.

\subsection{Training}

Let $\Theta$ denote all the parameters in the network which includes the weights associated with each of the layers described in previous sections. The entire network is then trained to minimize the following loss function:
\begin{equation}
  J(\Theta) = - \sum\limits_{c = 1}^C\mathbf{1}[y^*=c]\log\mathrm{Pr}(Y=c|\matr{x})
\end{equation}

where $C$ is the number of harm severity classes and y* is the ground truth label for the input report $\matr{x}$, $\mathbf{1}[\cdot]$ is the indicator function, and the probability of each harm severity class is estimated through the network.


\section{Experiments}
\label{sec:experiments}

\subsection{Data}
We use two large scale datasets consisting of patient safety and incidents reports sampled from various healthcare systems. These reports are sometimes referred to as ``patient safety reports'' in the health informatics literature, but in general they are meant to identify and characterize errors in patient care.


\begin{table}[t]
\centering
\caption{Dataset characteristics. Avg. refers to the average and std is the standard deviation.}
\label{tab:data1}
\small
\begin{tabular}{@{}lrr@{}}
\toprule
Statistic         & Dataset 1 & Dataset 2 \\ \midrule
Number of reports & 28,539          & 248,213    \\
Avg. report length (character)    & 411.4          & 239.2     \\
Stdev report length (character) & 370.9          & 187.2   \\ \bottomrule
\end{tabular}
\end{table}



\begin{table}[t]
\centering
\caption{Distribution of harm levels across different severity categories. Numbers show percentage of the entire data. The severity increases as we move from A to I. For the definition of the harm severity levels refer to Table \ref{tab:cat} at page 2.}
\label{tab:data-harm-levels}
\small
\setlength{\tabcolsep}{6pt}
\footnotesize
\begin{tabular}{@{}lrrrrrrrrrr@{}}
\toprule
      &  \multicolumn{4}{c}{no-harm}  & \multicolumn{5}{c}{harm} \\
 \cmidrule(lr){2-5} \cmidrule(lr){6-10}
Dataset      &  A    & B       & C      & D      & [E-I]   \\ \midrule
1       &39.3  &  13.8  & 19.9  & 13.1  & 13.9    \\
2       &11.7  &  12.9  & 40.4  & 31.6  & 3.4     \\
\bottomrule
\end{tabular}
\end{table}

This study was approved by the MedStar Health Research Institute Institutional Review Board (protocol 2014-101). The characteristics of the datasets are outlined in Table \ref{tab:data1}. We observe that one of the datasets (DS2) is larger than the other one (DS1) and the length of reports are rather different between them. Each dataset consists of reports regarding different categories in patient care.
The statistics about each of the harm levels are shown in Table \ref{tab:data-harm-levels}. The harm events (right side of the Table) are usually much less frequent than the events with no actual harm (left side of the Table). We divide each dataset to 3 subsets of training, validation, and test with respective distribution of 60\%, 20\%, and 20\% of the entire data. The hyper-parameters of the models were chosen empirically based on the performance on the validation set and the test set is preserved for evaluation.

\subsection{Evaluation}
We evaluate the effectiveness of our models in identifying harm by using standard classification evaluation metrics namely precision, recall, F-1 and Area Under the Curve (AUC).

\subsection{Baselines}
For comparing the performance of the proposed methods, we consider the following baselines:
\begin{itemize}[leftmargin=3mm]
  \item \I{SVM bow} - SVM with linear kernel with n-gram bag of words (bow) features \cite{wang2012baselines}. We experiment with three types of features, n-grams of size \{1\}, \{1,2\}, and \{1,2,3\} (we respectively abbreviate the resulting models with \I{bow1}, \I{bow2}, and \I{bow3}).
  \item \I{MNB bow} - We also experiment with Multinomial Naive Bayes method for classification where \citet{wang2012baselines} show its effectiveness in many text classification tasks. We used scikit-learn (http://scikit-learn.org/) implementation of SVM and MNB.
  \item \I{CNN} - We consider CNN model for text classification which has shown good results in both general domain \cite{kalchbrenner:2014,kim:2014} and biomedical domain \cite{rios2015convolutional}.
  \item \I{LSTM} - We also compare against RNN (LSTM) classifier which is similar to the models used in \cite{tang2015document,liu2016recurrent} (see Figure \ref{fig:rnn}).
\end{itemize}
  These methods form strong baselines with which we compare the performance of our models.

  \begin{table}[tb]
  \centering
  \caption{The results of identifying harm vs no-harm events. The top part of the Table shows the baselines while the bottom part shows the variants of the models presented in this work. Metrics are precision (P.), recall (R.), F-1 score for the harm category as well as the ROC Area Under the Curve (AUC). The numbers are percentages.}
  \label{tab:res-coarse}
  \setlength{\tabcolsep}{3.3pt}
  \footnotesize
  \begin{tabular}{@{}llllllllr@{}}
  \toprule
  \multirow{2}{*}{Method} & \multicolumn{4}{@{}c@{}}{Dataset 1} & \multicolumn{4}{@{}c@{}}{Dataset 2} \\
                  \cmidrule(l{5pt}r{10pt}){2-5} \cmidrule(l{5pt}r{10pt}){6-9}
            & P    & R    & F-1    &  AUC       & P      & R      & F-1    & AUC       \\ \midrule
            \multicolumn{9}{@{}l@{}}{\I{Baselines}} \\
  SVM bow1        & 81.5 & 52.4 & 63.8 & 89.2         & 85.1   & 61.7   & 71.5   & 94.0      \\
  SVM bow2        & 81.8 & 55.7  & 66.3 & 89.7         & 84.8   & 64.9   & 73.5   & 94.6      \\
  SVM bow3        & 81.9 & 55.1  & 65.9 & 89.7         & 84.5   & 65.6   & 73.9   & 94.8      \\
  MNB bow1      & 81.1 & 52.8  & 64.0 & 83.0         & 66.6   & 73.9   & 70.1   & 89.3      \\
  MNB bow2      & 86.6 & 43.1  & 57.5 & 79.0         & 73.9   & 66.0   & 69.7   & 86.9      \\
  MNB bow3      & \B{88.3} & 37.7  & 52.9 & 77.2         & 77.0   & 60.0   & 67.5   & 85.1      \\
  CNN             & 75.7 & 63.9  & 69.3 & 90.4        & 80.2   & 67.2   & 73.1   & 94.7      \\
  LSTM             & 76.6 & 61.9  & 68.8 & 90.3        & 70.1   & \B{75.9}   & 72.9   & 94.6      \\
  \midrule
            \multicolumn{9}{@{}l@{}}{\I{This work}} \\
  GRU CNN         & 75.8 & 68.3  & 71.8 & 91.0        & 77.9   & 73.2   & 75.5   & 95.0        \\
  Bi-GRU CNN      & 72.3 & 71.8  & 72.1 & 91.1        & 80.1   & 71.0   & 75.3   & 94.9      \\
  LSTM CNN        & 77.1 & 67.1  & 71.8 & 91.2        & 77.6   & 74.0   & 75.8   & 95.0        \\
  Bi-LSTM CNN     & 78.7 & 62.8  & 69.8 & 91.1        & 79.6   & 70.7   & 74.9   & 94.9      \\
  ATT GRU CNN     & 78.1 & 64.4  & 70.6 & 91.0        & 78.0   & 75.1   & 76.5   & 95.0        \\
  ATT Bi-GRU CNN  & 73.4 & 69.3  & 71.3 & 91.0        & \B{87.3}   & 70.3   & \B{77.9}   & 94.8      \\
  ATT LSTM CNN    & 69.4 & \B{76.8}  & \B{72.9} & \B{91.2}    & 78.8   & 74.9   & 76.8   & 95.0      \\
  ATT Bi-LSTM CNN & 83.0 & 64.0  & 72.3 & 91.0        & 79.9   & 74.5   & 77.1   & \B{95.2}        \\
  \bottomrule
  \end{tabular}
  \end{table}

\subsection{Model variants}
We evaluate several variants of our models. The first variant is our model architecture which is the entire model presented in \S \ref{sec:method} minus the attention model. We consider two types of recurrent networks, GRU and LSTM, as well as their bidirectional variants (Bi-GRU and Bi-LSTM). We then evaluate our complete model which utilizes the attention mechanism. When considering attention, we evaluated both GRU and LSTM as the underlying recurrent layer. We abbreviate these models based on the layers from top to bottom. For example ``\I{ATT GRU CNN}'' corresponds to our attention model with GRU unit, while example ``\I{ATT Bi-LSTM CNN}'' corresponds to our attention model with bidirectional LSTM in the recurrent layer.

\paragraph{Design decisions and hyperparameters} We empirically made the following design choices and hyperparameter selection: We used embedding size of 100 for word vectors and we set the maximum sequence length to 100 words (smaller sequences are padded with zero vectors and larger sequences are cropped). For convolution, we used filters of length 2 to 5 with 128 channels each, max pooling of length 4, and merge the output of the filters by concatenation. For RNN, we use LSTM and GRU with hidden size of 100. We used dropout rate of 0.25 after convolution. Training was done with batch size of 128 and through 2 and 6 epochs for the larger and smaller datasets, respectively. Adam \cite{kingma2014adam} was used as optimizer and early stopping was applied by monitoring accuracy on the validation set.

\begin{table}[]
\centering
\caption{The performance of our top model variants in identifying more fine grained types of harm. For the definition of the categories refer to \S \ref{sec:intro} and Table \ref{tab:cat}.
 The numbers are F-1 scores percentages in each category.}
\label{tab:fine-grained-4cat}
\vspace{-6pt}
\footnotesize
\setlength{\tabcolsep}{3.2pt}
\begin{tabular}{@{}lllllr@{}}
\toprule
& \begin{tabular}[c]{@{}l@{}}Temp./\\perm. harm\end{tabular} & \begin{tabular}[c]{@{}l@{}}Didn't \\ reach pt.\end{tabular} & \begin{tabular}[c]{@{}l@{}}Near\\ Miss\end{tabular} & Unsafe & Avg  \\ \midrule
\multicolumn{6}{@{}l}{\textit{Dataset1}}\\
  MNB bow2        & 41.4                                                                  & 39.7                                                               & 67.7                                                & 77.0   & 64.1 \\
  SVM bow2        & 53.5                                                                  & 49.8                                                               & 68.1                                                & 77.1   & 66.6 \\
  LSTM CNN        & 62.5                                                                  & 52.36                                                              & 67.7                                                & 75.8   & 68.2 \\
  CNN             & 64.0                                                                  & 47.4                                                               & 65.4                                                & 75.4   & 66.8 \\
  ATT Bi-LSTM CNN & 64.1                                                                  & 51.6                                                               & 68.5                                                & 76.1   & 68.7 \\
  ATT Bi-GRU CNN  & 64.5                                                                  & 48.9                                                               & 69.7                                                & 77.0   & \B{68.9} \\ \midrule
\multicolumn{6}{@{}l}{\textit{Dataset2}}\\
  MNB bow2        & 43.4                                                                  & 52.0                                                               & 82.8                                                & 51.8   & 71.0 \\
  SVM bow2        & 49.3                                                                  & 56.4                                                               & 82.1                                                & 55.7   & 72.6 \\
  LSTM CNN        & 62.9                                                                  & 55.1                                                               & 83.8                                                & 54.2   & 73.8 \\
  CNN             & 58.5                                                                  & 54.6                                                               & 82.7                                                & 55.0   & 72.6 \\
  ATT Bi-LSTM CNN & 59.3                                                                  & 59.2                                                               & 83.8                                                & 54.6   & 74.1 \\
  ATT Bi-GRU CNN  & 66.6                                                                  & 60.2                                                               & 83.4                                                & 59.3   & \B{75.3} \\
\bottomrule
\end{tabular}
\vspace{-6pt}
\end{table}

\subsection{Results}
We first consider the problem of identifying harm cases in the patient reports. That is, we classify a report as indicating some signs of harm to the patient (a harm case) or not (a no-harm case).
The main results of our methods in identifying harm are illustrated in Table \ref{tab:res-coarse}. The metrics are Precision, Recall, F-1 score for the harm category as well as the Area under the curve. We observe that our attention models (starting with ATT in the Table) are the best performing methods in both datasets evaluated by F-1 scores. In particular, the attention model using an LSTM recurrent unit (ATT LSTM CNN) achieves the highest F-1 of 72.9\% on the first dataset and the attention model using a bidirectional GRU (ATT Bi-GRU CNN) achieves F-1 of 77.9\% on the second dataset. While the results ranges are similar between the two datasets, in general we can see that the results on the second dataset are slightly higher. This is due to the datasets being generated at different healthcare systems and thus there are qualitative and quantitative difference between the datasets. As far as the baselines, we can see that in general, in terms of F-1 scores, traditional bag of words approaches \cite{wang2012baselines} are not quite competitive. In terms of precision, the Multinommial Naive Bayes method using up to 3gram features (MNB bow3) achieves the highest overall scores on first dataset; however, its recall is very low, making it relatively ineffective. The SVM baselines work generally better on the second dataset compared with the first dataset, and they outperform the performance of CNN and LSTM baselines. For example, the best F-1 score on the second dataset is 73.9\% which is for the SVM bow3 baseline. Our methods are still able to significantly improve over this baseline (compare the performance of \I{ATT Bi-GRU CNN} with \I{SVM bow 3}). Another trend that is worth noting is the significantly higher recall performance of our proposed models in comparison with the baselines. Recall is important in the task of harm detection, as any harm case can impact the patient and the method should minimize false negatives.
 We then compare the result of our method using a recurrent model on top of a convolutional model and observe how it can improve both the CNN and LSTM baselines. This suggests that while CNNs are effective in capturing the information in longer sequences, there is also some additional information that is captured when considering the interactions between the words along the entire sequence. We also observe that using a recurrent layer on top of a convoluational layer improves the performance (compare LSTM with our models in the Table), suggesting that local features captured by CNN are important in the final prediction.

Next, we evaluate the performance of our top models in fine-grain classification of harm severity on patients compared with the top baselines. Table \ref{tab:fine-grained-4cat} shows the performance on 4 levels of harm: Temporary or permenant harm, event that reached the patient but did not cause harm, near miss events, unsafe events. For description of these categories refer to Section \ref{sec:intro} and Table \ref{tab:cat}. We observe that our method variant \I{ATT Bi-GRU CNN} achieves the best overall performance with average respective F-1 scores of 68.9\% and 75.3\% in datasets 1 and 2.

\begin{table}[tb]
\centering
\caption{The results of our best method for the 1st dataset (ATT LSTM CNN) on different safety categories. We only show resuls for common categories.}
\label{tab:res-category-db1}
\setlength{\tabcolsep}{4.6pt}
\footnotesize
\begin{tabular}{@{}lrrrrr@{}}
\toprule
Category                         & F-1    &  P    &  R & Acc     \\ \midrule
Skin/Tissue                      & 92.9 &  94.1 & 91.8 & 87.2    \\
Surgery/Procedure                & 76.1 &  79.4 & 73   & 84.2    \\
Restraints/Seclusion Injury      & 75   &  75   & 75   & 90.9    \\
Airway Management                & 73.7 &  70   & 77.8 & 81      \\
Blood Bank                       & 71.4 & 100   & 55.6 & 98.1    \\
Lines/Tubes/Drain                & 66   &  70   & 62.5 & 71.7    \\
Medication/Fluid                 & 63.9 &  77.5 & 54.4 & 93      \\
Safety/Security                  & 56.4 &  75.9 & 44.9 & 73.2    \\
Diagnosis/Treatment              & 55   &  56.9 & 53.2 & 79.3    \\
Miscellaneous                    & 55   &  54.5 & 55.6 & 83.7    \\
Fall                             & 52.2 &  63.8 & 44.1 & 86.4    \\
Diagnostic Imaging               & 50   &  44   & 57.9 & 83.7    \\
Patient ID/Documentation & 36.4 &  40   & 33.3 & 97      \\
Lab/Specimen                     &  0   &   0   &  0   & 97.2    \\
\bottomrule
\end{tabular}
\end{table}

\begin{table}[tb]
\centering
\caption{The results of harm identification for our best method on the second dataset (ATT Bi-GRU CNN) on different safety categories. Numbers are percentages. P/T/T refers to the category of Procedure/Treatment/Test}
\label{tab:res-category-db2}
\setlength{\tabcolsep}{4.6pt}
\footnotesize
\begin{tabular}{@{}lrrrrrr@{}}
\toprule
Category                         & F-1    &  P    &  R & Acc   \\ \midrule
Complication of P/T/T      & 81.8 & 80.6 & 83.1 & 85.7  \\
Fall                       & 79.7 & 86   & 74.3 & 95    \\
Error in P/T/T             & 71.3 & 68.8 & 74   & 96.3  \\
Miscellaneous              & 68.4 & 68.2 & 68.5 & 87     \\
Skin Integrity             & 66.4 & 75.2 & 59.4 & 92.2  \\
Equipment/Supplies/Devices & 61.7 & 62.5 & 61   & 95.9   \\
Transfusion                & 52.4 & 68.8 & 42.3 & 96     \\
Medication error           & 49.4 & 69.4 & 38.3 & 98.3  \\
Adverse Drug Reaction      & 46.2 & 61.5 & 37   & 80.8   \\
\bottomrule
\end{tabular}
\vspace{6pt}
\end{table}

\begin{figure}[b]
\begin{center}
\includegraphics[scale=0.20]{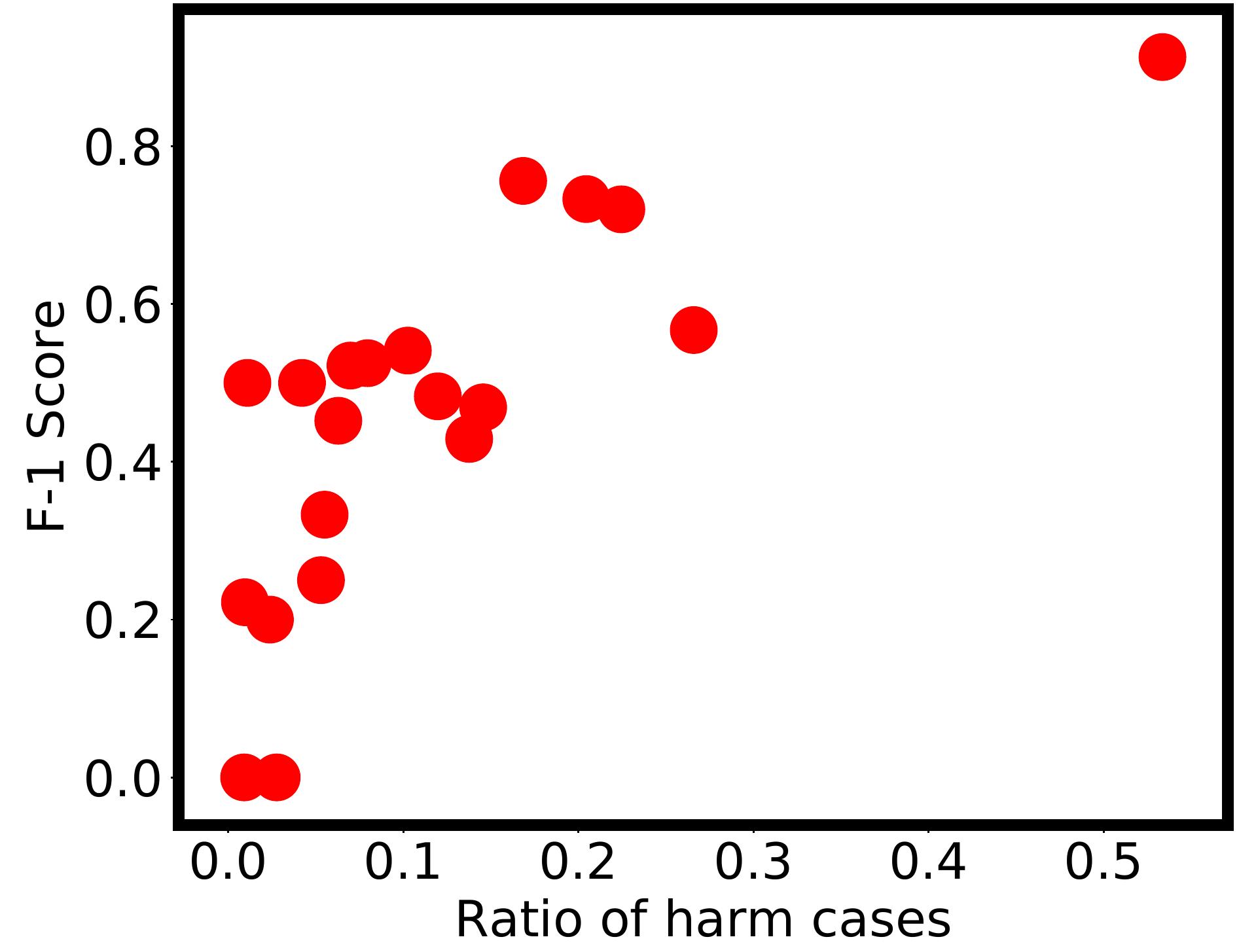}
\caption{The performance of the our best method on dataset 1 based on each category. Each data point shows the results in a specific category as well as the ratio of harm cases in that category. The x-axis shows the ratio of harm cases.}
\label{fig:res-cat}
\end{center}
\end{figure}

\subsection{Analysis}
To better evaluate the performance of our system and study the errors that it makes, we analyze the performance on each dataset based on each category of incident reports. The incident reports are categorized into several categories and there are often qualitative differences between the narratives in different categories. Tables \ref{tab:res-category-db1} and \ref{tab:res-category-db2} show the breakdown of results based on top common categories in dataset 1 and 2, respectively. We report the results of the best performing model variant on each dataset (i.e. \textit{ATT LSTM CNN} for dataset 1 and \textit{ATT Bi-GRU CNN} for dataset 2). On dataset 1 (Table \ref{tab:res-category-db1}) we observe that the model achieves very high scores in identifying harm in Skin/Tissue category with F-1 of 92.9\% in identifying harm. Results on some other categories such as \I{Surgury/Procedure}, \I{Seclusion Injury}, \I{Airway Management}, and \I{Blood bank} are also relatively high. However, we observe that on some categories such as \I{Patient ID/Documentation} and \I{Lab/Specimen} the performance is low. We attribute the low performance in these categories to three main reasons: the total number of data in each category, the relative number of harm cases in the category, and the diversity of the type of reports in each category.
 We analyzed the distribution of the harm cases in each category. Some categories are more balanced in terms of harm and no-harm cases, while other categories are extremely unbalanced. We calculate the class ratios of harm in each category and compare the results based on these ratios. Figure \ref{fig:res-cat} illustrates the performance of our method on each category and the ratio of harm cases in that category. Each data point shows the performance results in terms of F-1 based on the ratio of harm cases in that category. We observe that as the ratio of harm cases increases, the performance generally tends to increase. This is expected, as training the model on highly unbalanced datasets prevents the model to learn the appropriate weights associated with the positive class. The two categories at the bottom left side of Figure \ref{fig:res-cat} are the categories with lowest results in Table \ref{tab:res-category-db1}. The respective ratio of harm cases in these categories are 0.009 and 0.027, while the ratio of harm cases in \I{Skin/Tissue} (the point on top right side of the Figure) is 0.53.

We also performed qualitative analysis on the reports in each category by inspecting the type of incidents in each category. We investigate the types of incidents in the best performing category in dataset 1 \I{Skin/Tissue} and most of the events are regarding pressure ulcer and wounds.
On the other hand, looking at the \I{Lab/Specimen} category, there are many diverse types of errors and harm in this category such as collection issues, documentation problems, labeling issues, ordering issues, etc, that are very different in description, making it difficult for the model to learn all the nuances in this category.
 This reason, coupled with relative low number of harm cases in the dataset in this category, results in low performance. We believe that having more data would help improving the performance of the model.

\section{Conclusions}
In this paper, we presented a neural network model for identifying harm in safety narratives related to clinical care. We used a multi-layer network with convolutional, recurrent, and soft attention mechanism layers. We argued that convolutional layer is important in finding the local features and the recurrent layer with attention is effective in finding the interactions and dependencies along the sequence.
 We demonstrated that our methods can significantly improve the performance over existing methods in identifying harm safety cases. The impact of the methods and results presented in this paper is substantial to patient care. More accurate methods in the identification of harm can help the data analysis and reporting process, prevent harm to patients, better prioritize resources to address safety incidents, and subsequently improve general patient care.

\section*{Acknowledgments}
We thank the four anonymous reviewers for their helpful comments and suggestions. We gratefully acknowledge the support of NVIDIA Corporation with the donation of the Titan X Pascal GPU used for this research. This project was funded under contract/grant number Grant R01 HS023701-02 from the Agency for Healthcare Research and Quality (AHRQ), U.S. Department of Health and Human Services. The opinions expressed in this document are those of the authors and do not necessarily reflect the official position of AHRQ or the U.S. Department of Health and Human Services.

\bibliographystyle{ACM-Reference-Format}
\bibliography{sigconf}

\end{document}